\documentclass[letterpaper]{article} 
\usepackage{aaai2026}  
\usepackage{times}  
\usepackage{helvet}  
\usepackage{courier}  
\usepackage[hyphens]{url}  
\usepackage{graphicx} 
\usepackage{natbib}  
\usepackage{caption} 
\usepackage{algorithm}
\usepackage{algorithmic}
\usepackage{booktabs}
\usepackage{pifont}
\usepackage{amsmath}
\usepackage{multirow}
\usepackage{diagbox}
\usepackage{makecell}
\usepackage{xcolor}

\usepackage{newfloat}
\usepackage{listings}
\DeclareCaptionStyle{ruled}{labelfont=normalfont,labelsep=colon,strut=off} 
\floatstyle{ruled}
\newfloat{listing}{tb}{lst}{}
\floatname{listing}{Listing}
\nocopyright 

\title{MME-SCI: A Comprehensive and Challenging Science Benchmark for Multimodal Large Language Models}
\author{
    Jiacheng Ruan\textsuperscript{\rm 1}\equalcontrib,
    Dan Jiang\textsuperscript{\rm 1}\equalcontrib, 
    Xian Gao\textsuperscript{\rm 1},
    Ting Liu\textsuperscript{\rm 1}, 
    Yuzhuo Fu\textsuperscript{\rm 1\dag},
    Yangyang Kang\textsuperscript{\rm 2}\thanks{Yuzhuo Fu and Yangyang Kang are co-corresponding authors.}
}
\affiliations{
    \textsuperscript{\rm 1} Shanghai Jiao Tong University,
    \textsuperscript{\rm 2} Bytedance
    \\
    jackchenruan@sjtu.edu.cn

}

\usepackage{bibentry}

\begin{document}

\maketitle

\begin{abstract}

Recently, multimodal large language models (MLLMs) have achieved significant advancements across various domains, and corresponding evaluation benchmarks have been continuously refined and improved. In this process, benchmarks in the scientific domain have played an important role in assessing the reasoning capabilities of MLLMs. However, existing benchmarks still face three key challenges: \textbf{1)} Insufficient evaluation of models' reasoning abilities in multilingual scenarios; \textbf{2)} Inadequate assessment of MLLMs' comprehensive modality coverage; \textbf{3)} Lack of fine-grained annotation of scientific knowledge points. To address these gaps, we propose MME-SCI, a comprehensive and challenging benchmark. We carefully collected 1,019 high-quality question-answer pairs, which involve 3 distinct evaluation modes. These pairs cover four subjects, namely mathematics, physics, chemistry, and biology, and support five languages: Chinese, English, French, Spanish, and Japanese. We conducted extensive experiments on 16 open-source models and 4 closed-source models, and the results demonstrate that MME-SCI is widely challenging for existing MLLMs. For instance, under the Image-only evaluation mode, o4-mini achieved accuracy of only 52.11\%, 24.73\%, 36.57\%, and 29.80\% in mathematics, physics, chemistry, and biology, respectively, indicating a significantly higher difficulty level compared to existing benchmarks. More importantly, using MME-SCI's multilingual and fine-grained knowledge attributes, we analyzed existing models' performance in depth and identified their weaknesses in specific domains. For example, in questions related to ``Magnetic Field'', o4-mini correctly answered only 5 out of 33 questions, thereby fine-grainedly exposing the model's vulnerabilities. These findings highlight the urgent need to enhance the scientific reasoning capabilities of MLLMs. 
The Data and Evaluation Code are available at https://github.com/JCruan519/MME-SCI.

\end{abstract}
\section{Introduction}

\begin{figure}[!t]
    \centering
    \includegraphics[width=0.99\linewidth]{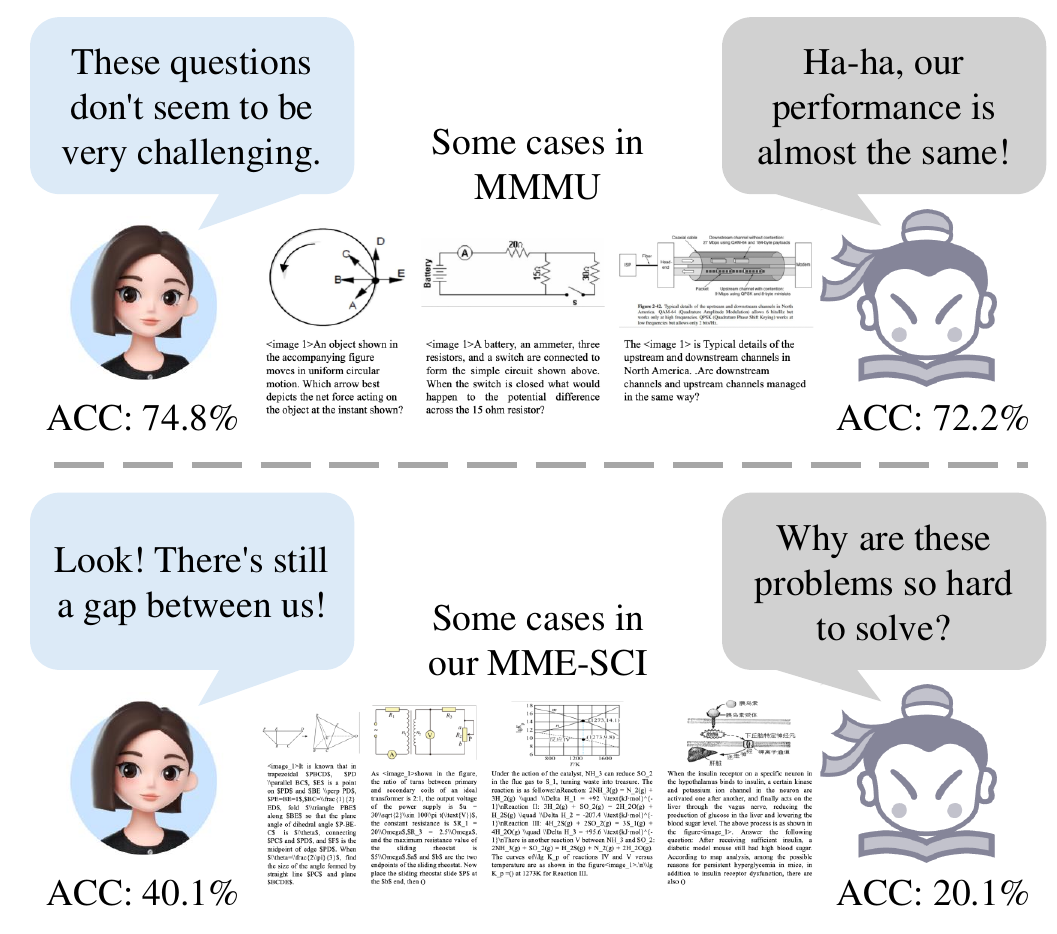}
    \caption{Comparison between prevalent benchmarks (e.g., MMMU) and MME-SCI. Existing benchmarks for MLLMs have become saturated and fail to distinguish performance differences among models. ACC stands for accuracy.}
    \label{fig:head}
\end{figure}

In recent years, Multimodal Large Language Models (MLLMs) have achieved breakthrough progress in tasks such as visual question answering and multimodal reasoning \cite{mllm_reasoning_survey_1,mllm_reasoning_survey_2,mllm_reasoning_survey_3}, with representative models including the GPT series \cite{gpt4,o4mini}, Qwen-VL series \cite{Qwen2-vl,Qwen2.5VL}, and InternVL series \cite{InternVL2.5,internvl3}. However, as shown in Figure \ref{fig:head}, compared with the rapid improvement of model capabilities, existing multimodal evaluation benchmarks—especially those for scientific-related scenarios—have shown a trend of ``being unable to keep up''. Specifically, on MMMU \cite{mmmu}, a comprehensive benchmark based on university multidisciplinary problems, InternVL3-78B has achieved an accuracy of 72.2\%, and Qwen2.5-VL-72B has also reached 68.2\%. On AI2D \cite{ai2d}, a benchmark for scientific chart, even smaller models such as InternVL3-8B and Qwen2.5-VL-7B have achieved accuracies of 85.1\% and 84.3\%, respectively. These results indicate that the performance of advanced MLLMs on existing mainstream scientific benchmarks has approached a saturation level. Thus, there is an urgent need to design a challenging benchmark to continuously drive breakthroughs in models' scientific reasoning and cross-modal understanding capabilities.

A high-quality scientific evaluation benchmark is the key premise for accurately assessing the reasoning capabilities of MLLMs in scientific domains. The design of such a benchmark should meet the following three core characteristics: 
\textbf{1) Multilingual adaptability.} Currently, the training corpus for MLLMs is predominantly in English, leading to performance differences when the same model is applied to non-English contexts. More importantly, multilingual scenarios are more effective in verifying whether MLLMs have truly mastered reasoning abilities, rather than relying on specific linguistic contexts. Furthermore, Cross-linguistic consistency is essential for supporting global scientific collaboration.
\textbf{2) Comprehensive modality coverage.} Most existing benchmarks tend to focus on image-text hybrid scenarios and lack systematic testing of MLLMs on image-only and text-only modes. Evaluations with diverse modalities are essential to reflect the robustness and applicability of MLLMs in real-world applications. 
\textbf{3) Fine-grained knowledge points annotation.} Current benchmarks are still deficient in the systematic annotation of knowledge points, making it difficult for evaluation results to provide targeted feedback and thereby limiting in-depth analysis of models' potential flaws and disciplinary weaknesses.

To address the aforementioned gap, we introduce MME-SCI, a comprehensive and highly challenging multimodal evaluation benchmark. This benchmark consists of 1,019 high-quality question-answer pairs that have undergone manual selection, covering three question types: single-choice, multiple-choice, and fill-in-the-blank. The content spans four core subjects: mathematics, physics, chemistry, and biology. Additionally, it provides five language versions (e.g., Chinese, English, French, Spanish, and Japanese) to support multilingual evaluation. Furthermore, we have annotated each question with the corresponding knowledge point, covering a total of 63 fine-grained concepts, and designed three input modalities: text-only, image-only, and image-text hybrid. These designs collectively offer a comprehensive and detailed evaluation framework for systematically assessing the reasoning capabilities of MLLMs across diverse linguistic environments and knowledge systems.

\begin{table}[!t]
\centering\footnotesize
\setlength{\tabcolsep}{2pt}
\begin{tabular}{lcccc}
\toprule
\textbf{Benchmark} &  \textbf{ML.} & \textbf{CMC.} & \textbf{MD.} &\textbf{FKP.} \\
\midrule
GAOKAO-Bench \cite{gaokaobench} &\ding{51} &\ding{55}  &\ding{51} &\ding{55} \\
MathVerse \cite{mathverse}   &\ding{55} &\ding{51}  &\ding{55} &\ding{51} \\
MATH-Vision \cite{mathvision} &\ding{55} &\ding{55}  &\ding{55} &\ding{51} \\
MMMU \cite{mmmu}       &\ding{55} &\ding{55}  &\ding{51} &\ding{51} \\
EMMA \cite{emma}        &\ding{55} &\ding{55}  &\ding{51} &\ding{51} \\
GeoSense \cite{geosense}    &\ding{51} &\ding{55}  &\ding{55} &\ding{51} \\
PhyX \cite{PhyX}    &\ding{55} &\ding{55}  &\ding{55} &\ding{51} \\
VisioMath \cite{visiomath}    &\ding{55} &\ding{55}  &\ding{55} &\ding{55} \\
\midrule
\textbf{MME-SCI} (Ours)  &\ding{51} &\ding{51}  &\ding{51} &\ding{51} \\
\bottomrule
\end{tabular}
\caption{Comparison between our MME-SCI and others. \textbf{ML.} denotes Multilingual, \textbf{CMC.} signifies Comprehensive Modality Coverage, \textbf{MD.} represents Multidisciplinary, and \textbf{FKP.} stands for Fine-grained Knowledge Points.}
\label{tab:comparison_bmk}
\end{table}

We conducted comprehensive experiments of 16 open-source and 4 closed-source models on the MME-SCI benchmark. Taking the Image-only evaluation mode as an example: Qwen2.5VL-72B, the top-performing open-source model, achieved an accuracy of only 19.43\%, while Doubao-Seed-1.6 \cite{doubao1.6}, an advanced representative among closed-source models, reached an accuracy of merely 41.32\%. These results demonstrate that MME-SCI poses significant challenges to existing MLLMs. Furthermore, leveraging the advantages of MME-SCI, such as multilingual support and fine-grained knowledge points, we are able to conduct in-depth analysis of the performance of existing models, thereby accurately revealing their shortcomings in language consistency and domain-specific applications.

The main contributions of this paper could be summarized as follows: 
\begin{itemize}
    \item We propose MME-SCI, a comprehensive multimodal scientific benchmark. This benchmark contains 1,019 manually curated samples, supports 5 languages and 3 input modalities, and covers 63 knowledge concepts in mathematics, physics, chemistry, and biology. It effectively addresses the limitations of existing benchmarks in multilingual adaptability, comprehensive modality coverage, and fine-grained knowledge annotation.
    \item Evaluation results based on 20 MLLMs show that MME-SCI poses significant challenges to current models and breaks the performance saturation of existing mainstream scientific evaluation benchmarks.
    \item Leveraging the strengths of MME-SCI in multilingual support and fine-grained annotation, we perform in-depth analyses and accurately identify the limitations of models in cross-lingual consistency, modality-specific reasoning, and discipline-specific knowledge, thereby offering targeted guidance for improving MLLMs.
\end{itemize}
\section{Related works}

\subsection{Multimodal Large Language Models}

With the rapid advancement of LLMs and visual foundation models, MLLMs have demonstrated remarkable breakthroughs in multimodal tasks. The open-source community has taken the lead in promoting technological implementation through lightweight solutions: LLaVA \cite{llava} freezes the CLIP \cite{clip} for image encoding and injects visual prompts into the LLM decoder \cite{vicuna} via a lightweight projection layer to achieve cross-modal alignment; the Qwen-VL series \cite{qwenvl,Qwen2-vl,Qwen2.5VL} enhances spatiotemporal perception by progressively upgrading visual encoders, introducing dynamic resolution mechanisms, and multimodal rotational position encoding, while expanding training data scale to enable the evolution from single-image understanding to unified image-video processing; the InternVL series \cite{internvl,internvl1_5,InternVL2.5,internvl3} has transitioned from the initially complex architecture based on BLIP-2 \cite{blip2} improvements to the simple `ViT-MLP-LLM' framework, achieving performance close to closed-source models like GPT-4V \cite{gpt4v} through gradual upgrades in model scaling, data optimization, and inference strategies during testing.

\begin{figure*}[!t]
    \centering
    \includegraphics[width=0.99\linewidth]{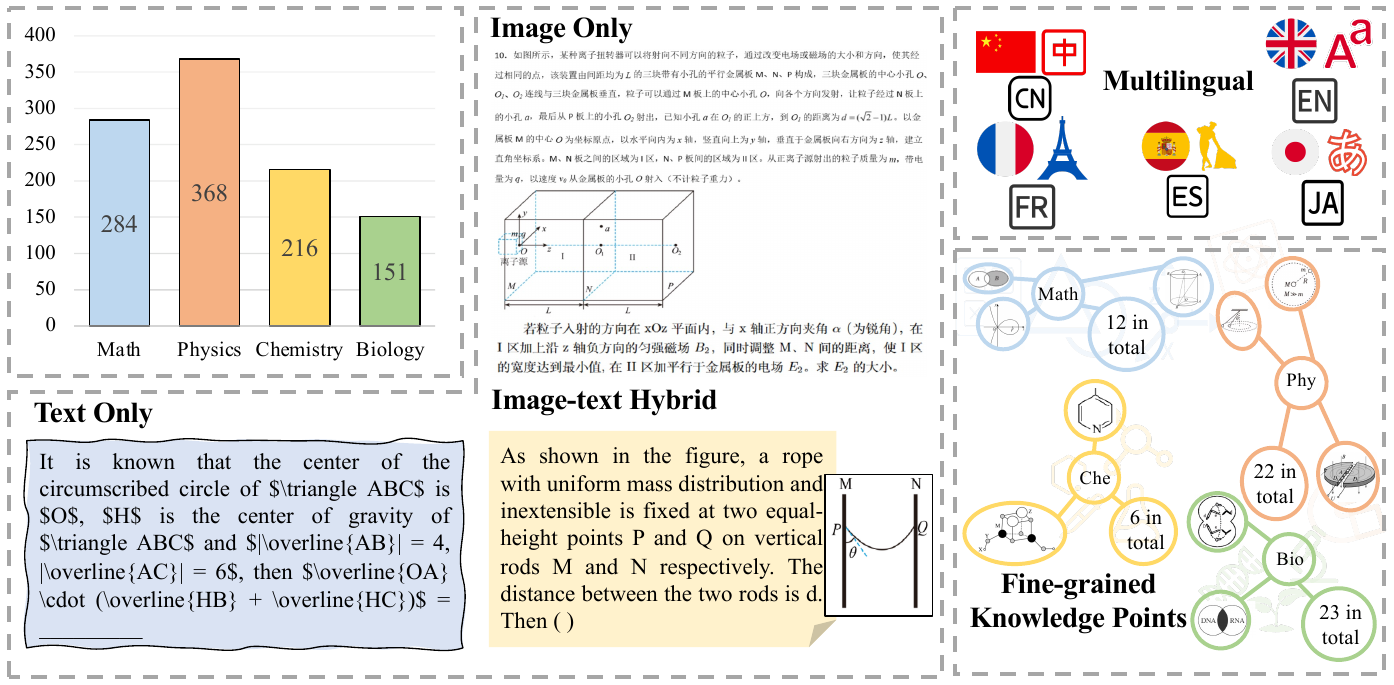}
    \caption{Overview of MME-SCI. This benchmark consists of 1,019 manually and carefully selected questions, covering four subjects, with the characteristics of multilingual support, full-modal coverage, and fine-grained knowledge points.}
    \label{fig:mmesci}
\end{figure*}

\subsection{Science-related Benchmarks}

With the continuous evolution of MLLMs capabilities, developing benchmarks that match their performance has become a critical research \cite{mllm_bmk_survey_1,mllm_bmk_survey_2}. Benchmarks in the multimodal field can be mainly categorized into natural domain and scientific domain: the former focuses on evaluating MLLMs' ability to understand natural images \cite{xlrsbench,Gmai-mmbench,Drivemllm}, while the latter centers on reasoning tasks in scientific disciplines. Specifically, MMMU \cite{mmmu} is a benchmark covering 6 disciplines and 30 courses, containing 11.5K university-level questions; EMMA \cite{emma} is an enhanced multimodal reasoning benchmark that includes questions in mathematics, physics, chemistry, and coding, aiming to assess models' visual reasoning abilities; GeoSense \cite{geosense} is a bilingual benchmark that evaluates MLLMs' geometric reasoning capabilities through a five-level hierarchical framework based on geometric principles and 1,789 questions. In addition, other multimodal benchmarks \cite{mmecot,mmreason,rbench} have also promoted the development of MLLMs. However, with the rapid iteration of MLLMs, these benchmarks have gradually become saturated. As shown in Table \ref{tab:comparison_bmk}, their core limitation lies in that existing benchmarks cannot fully cover the multi-dimensional capabilities of MLLMs. In contrast, our MME-SCI contains 1,019 manually curated question-answer pairs, simultaneously featuring characteristics such as Multilingual, Comprehensive modality coverage, Multidisciplinary, and Fine-grained knowledge points. This not only poses more rigorous challenges to existing MLLMs but also enables more dimensional performance analysis.

\begin{figure*}[!t]
    \centering
    \includegraphics[width=0.99\linewidth]{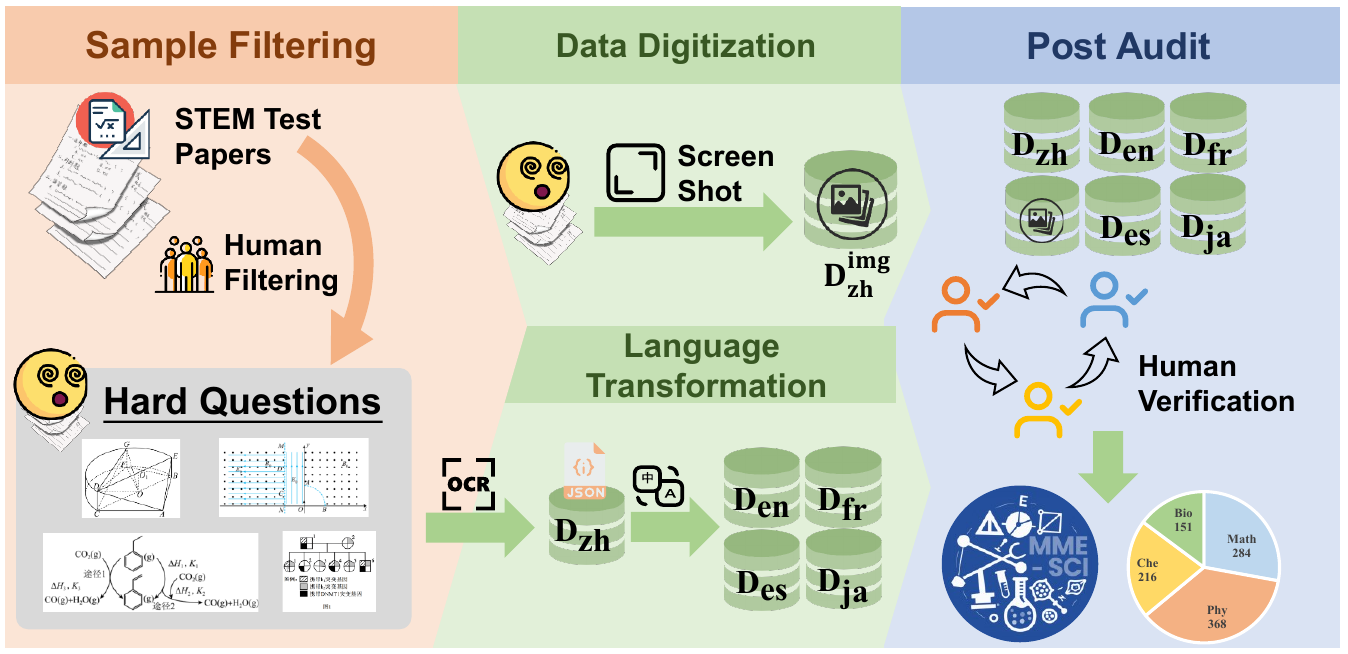}
    \caption{Construction pipeline of MME-SCI, which consists of three stages. Overall, a total of approximately 300 person-days were spent on problem selection, digitization, language conversion, and post-verification.}
    \label{fig:pipline}
\end{figure*}

\section{MME-SCI}

\subsection{Overview of MME-SCI}

As illustrated in Figure \ref{fig:mmesci}, we introduce MME-SCI, a comprehensive and challenging benchmark for scientific domain. It comprises 1,019 manually curated questions, systematically covering four subjects (mathematics, physics, chemistry, and biology) at the Chinese high school level. These four subjects contain 12, 22, 6, and 23 knowledge points, respectively, based on high school textbook systems and suggestions from subject experts. This enables MLLMs to accurately identify their weaknesses in specific knowledge points. In terms of question timeliness, 83.3\% of the questions are sourced from 2025, 16.2\% from 2024, and the remaining 0.5\% from years prior to 2024, ensuring the novelty of the data sources.

Given that the training corpora of current MLLMs are mostly centered on English scenarios, MME-SCI is specifically equipped with 5 language versions, including Chinese, English, French, Spanish, and Japanese, to comprehensively evaluate the cross-lingual reasoning capabilities of models. In terms of modal design, MME-SCI encompasses three evaluation modes: text-only, image-only and image-text hybrid. Specifically, text-only mode are designed to assess language comprehension abilities, image-only mode emphasize the evaluation of visual semantic parsing capabilities, and image-text mode require the integration of visual information and textual logic. These tasks collectively form a multi-dimensional evaluation system that covers full-modal reasoning abilities.

To reduce the complexity of evaluation, all questions are designed as single/multiple-choice or fill-in-the-blank items, paired with concise and verifiable answers. This design not only supports automated scoring but also facilitates manual verification. By incorporating multilingual scenario design, full-modal evaluation, and fine-grained knowledge points coverage, MME-SCI aims to serve as a core benchmark for systematically enhancing MLLMs' complex reasoning capabilities in scientific domains.

\subsection{Data Curation Process}

As illustrated in Figure \ref{fig:pipline}, the construction pipeline of MME-SCI consists of four core steps, namely sample filtering, data digitization, language transformation, and post-auditing, which are specified as follows:

\subsubsection{Sample Filtering}
We recruited 3 evaluation volunteers (2 senior undergraduates and 1 graduate student), all of whom ranked within the top 0.1\% in terms of average scores in China's College Entrance Examination (Gaokao). The volunteers were asked to solve questions from mock exam papers of high school science subjects\footnote{Mock high school exam papers were chosen instead of Gaokao papers because Gaokao papers have a wide dissemination range and low accessibility, making them more likely to be included in the training data of MLLMs.}, and to filter out the questions they answered incorrectly or found confusing.

\subsubsection{Data Digitization and Language Transformation}
After manually selecting high-difficulty questions, we recruited five annotators to process the data. Specifically, we used GPT-4o and OCR tools to extract the questions and answers, converting them into JSON format to form the $\boldsymbol{\mathrm{D_{zh}}}$. $\boldsymbol{\mathrm{D_{zh}}}$ contains 805 questions that require image-based understanding (such as problems in analytical geometry and circuit analysis) and 214 text-only questions. Subsequently, we took screenshots of all the questions in $\boldsymbol{\mathrm{D_{zh}}}$, creating the $\boldsymbol{\mathrm{D_{zh}^{img}}}$ for use in the image-only evaluation mode. Examples of this can be found in Figure \ref{fig:mmesci}, with more samples provided in the Appendix E. We then translated the Chinese content in $\boldsymbol{\mathrm{D_{zh}}}$ into English, French, Spanish, and Japanese, resulting in $\boldsymbol{\mathrm{D_{en}}}$, $\boldsymbol{\mathrm{D_{fr}}}$, $\boldsymbol{\mathrm{D_{es}}}$, and $\boldsymbol{\mathrm{D_{ja}}}$, respectively.

\subsubsection{Post-Audit}
In the final stage, we recruited three reviewers to perform cross-validation on the OCR results in $\boldsymbol{\mathrm{D_{zh}}}$, the integrity of the screenshots in $\boldsymbol{\mathrm{D_{zh}^{img}}}$, and the language conversion results. If one reviewer identifies an error, the other two will conduct a secondary verification. In the secondary verification process, if any reviewer identifies an error, a revision will be made to ensure the quality of the data.

\begin{table*}[!t]
\centering
\setlength{\tabcolsep}{2.5pt}
\renewcommand{\arraystretch}{1.0} 
\begin{tabular}{l|ccccc|ccccc|cccc}
\toprule
\multirow{2}{*}{\textbf{Model}} & \multicolumn{5}{c|}{\textbf{$\boldsymbol{\mathrm{D_{zh}}}$}} & \multicolumn{5}{c|}{\textbf{$\boldsymbol{\mathrm{D_{zh}^{img}}}$}} & {\textbf{$\boldsymbol{\mathrm{D_{en}}}$}} & {\textbf{$\boldsymbol{\mathrm{D_{fr}}}$}} & {\textbf{$\boldsymbol{\mathrm{D_{es}}}$}} & {\textbf{$\boldsymbol{\mathrm{D_{ja}}}$}} \\ 
\cline{2-15}
 & \textbf{math} & \textbf{phy} & \textbf{chem} & \textbf{bio} & \textbf{AVG.} & \textbf{math} & \textbf{phy} & \textbf{chem} & \textbf{bio} & \textbf{AVG.} & \textbf{AVG.} & \textbf{AVG.} & \textbf{AVG.} & \textbf{AVG.} \\ 

\midrule
\multicolumn{15}{l}{\textbf{\emph{Open-source LVLMs} $<$ 10B (Small group)}}\\
\hline

InternVL3-2B & 9.86 & 6.52 & 18.06 & 22.52 & 12.27  & 9.15 & 5.71 & 10.65 & 14.57 & 9.03  & 13.05 & 10.89 & 11.78 & 15.70 \\ 
Qwen2.5VL-3B & 10.56 & 7.88 & 18.98 & 25.17 & 13.54  & 8.10 & 7.61 & 14.35 & 21.85 & 11.29  & 16.68 & 14.03 & 14.62 & 11.58 \\ 
InternVL3-8B & 10.56 & 9.78 & 21.76 & 25.83 & 14.92  & 4.93 & 7.07 & 15.74 & 17.88 & 9.91  & 16.78 & 15.41 & 16.39 & 16.68 \\ 
Qwen2.5VL-7B & 8.80 & 11.41 & 22.69 & 25.83 & 15.21  & 10.21 & 13.32 & 23.61 & 27.15 & 16.68  & 15.80 & 15.80 & 15.21 & 15.70 \\ 
Qwen2.5-Omni-7B & 11.62 & 6.52 & 20.83 & 27.81 & 14.13  & 4.23 & 8.42 & 20.37 & 23.18 & 11.97  & 16.00 & 16.19 & 16.88 & 14.43 \\
Ovis2-8B & 15.14 & 8.15 & 18.06 & 27.81 & 15.11  & 10.92 & 6.52 & 12.96 & 18.54 & 10.89  & 16.00 & 14.72 & 15.11 & 10.89 \\ 
Llava-OneVision-7B & 8.10 & 7.88 & 15.74 & 20.53 & 11.48  & 2.46 & 12.23 & 6.94 & 17.22 & 9.13  & 13.74 & 13.64 & 13.54 & 10.30 \\ 
Kimi-VL-A3B-Instruct & 13.73 & 7.61 & 16.67 & 24.50 & 13.74  & 8.80 & 4.89 & 13.43 & 22.52 & 10.40  & 12.86 & 9.32 & 11.19 & 11.09 \\
Kimi-VL-A3B-Thinking & 23.94 & 10.33 & 22.22 & 35.76 & 20.41  & 22.97 & 8.97 & 18.98 & 29.14 & 17.98  & 21.10 & 22.18 & 21.59 & 13.25 \\ 
Phi-4-multimodal-instruct & 7.75 & 7.61 & 8.33 & 15.23 & 8.93  & 3.17 & 4.35 & 3.24 & 9.27 & 4.51  & 11.68 & 10.11 & 11.87 & 11.97 \\ 

\textbf{Avg. Performance} &12.01 & 8.37 & 18.33 & 25.10 & 13.97 & 8.49 & 7.91 & 14.03 & 20.13 & 11.18 & 15.37 & 14.23 & 14.82 & 13.16  \\ 

\midrule
\multicolumn{15}{l}{\textbf{10B $<$ \emph{Open-source LVLMs} $<$ 40B (Middle group)}}\\
\hline

Ovis2-16B & 15.85 & 7.88 & 20.83 & 28.48 & 15.90  & 9.86 & 6.52 & 18.06 & 19.87 & 11.87  & 17.08 & 15.01 & 16.09 & 15.11 \\ 
Ovis2-34B & 11.97 & 10.33 & 21.76 & 30.46 & 16.19 & 10.21 & 7.07 & 16.20 & 20.53 & 11.87  & 17.57 & 15.70 & 17.08 & 14.33 \\ 
Skywork-R1V-38B & 22.18  & 10.87  & 25.93  & 26.49  & 19.53  & 15.14 & 8.15 & 18.52 & 25.17 & 14.82  & 22.18  & 23.55 & 20.90 & 16.29 \\ 

\textbf{Avg. Performance} &16.67 & 9.69 & 22.84 & 28.48 & 17.21 & 11.74 & 7.25 & 17.59 & 21.86 & 12.85 & 18.94 & 18.09 & 18.02 & 15.24  \\ 

\midrule
\multicolumn{15}{l}{\textbf{\emph{Open-source LVLMs} $>$ 40B (Large group)}}\\
\hline

Llava-OneVision-72B & 13.93 & 10.35 & 20.83 & 32.45 & 16.86  & 4.58 & 7.69 & 9.30 & 11.41 & 7.71  & 18.84 & 14.92 & 16.98 & 16.88 \\ 
Qwen2.5VL-72B & 19.01 & 14.40 & 30.56 & 34.44 & 22.08  & 14.79 & 11.96 & 29.17 & 32.45 & 19.43  & 19.17 & 18.65 & 19.53 & 17.47 \\ 
InternVL3-78B & 16.25 & 11.72 & 26.39 & 27.15 & 18.39  & 15.30 & 12.98 & 21.30 & 25.83 & 17.33  & 20.12 & 20.12 & 19.45 & 19.43 \\ 

\textbf{Avg. Performance} &16.40 & 12.16 & 25.93 & 31.35 & 19.11 & 11.56 & 10.88 & 19.92 & 23.23 & 14.82 & 19.38 & 17.90 & 18.65 & 17.93   \\ 

\midrule
\hline

\multicolumn{15}{l}{\textbf{\emph{Closed-source LVLMs}}}\\

\hline

Claude-4-Sonnet & 21.13  & 17.12  & 31.48  & 33.77  & 23.75  & 19.43  & 13.59  & 23.61  & 21.19  & 18.47  & 22.96 & 22.28 & 24.14 & 22.37 \\ 
Doubao-1.5-tv-pro & 28.87 & 28.80 & 40.74 & 41.72 & 33.27 & 38.73  & 31.52  & 44.44  & 46.36  & 38.47  & 34.05 & 29.93 & 32.09 & 34.54 \\ 
Doubao-Seed-1.6 & 44.37 & 31.52 & 42.13 & 41.72 & 38.86 & 49.30  & 33.15  & 42.13  & 45.03   & 41.32  & 40.14 & 37.88 & 38.76 & 36.41 \\ 
o4-mini-20250416 & 50.35  & 26.36  & 38.89  & 35.10  & 37.00  & 52.11  & 24.73  & 36.57  & 29.80  & 35.62  & 33.17  & 29.54  & 31.11  & 29.64  \\

\textbf{Avg. Performance} &36.18 & 25.95 & 38.31 & 38.08 & 33.22 & 39.89 & 25.75 & 36.69 & 35.59 & 33.47 & 32.58 & 29.91 & 31.53 & 30.74  \\

\bottomrule
\end{tabular}
\caption{Results on MME-SCI. We present detailed comparative results for the $\boldsymbol{\mathrm{D_{zh}}}$ and $\boldsymbol{\mathrm{D_{zh}^{img}}}$ scenarios, including accuracy for four subjects. For the remaining four languages, we report the average accuracy (AVG.).}
\label{tab:main_res}
\end{table*}

\subsection{Differences from the Existing Benchmark}

Compared with existing benchmarks, MME-SCI exhibits significant advantages in multilingual adaptability, full-modal coverage, and fine-grained annotation of knowledge points. Existing benchmarks are mostly limited to a single language environment (e.g., MMMU \cite{mmmu} only supports English, while GAOKAO-Bench \cite{gaokaobench} is primarily in Chinese), making it difficult to evaluate models' reasoning capabilities in cross-lingual scientific scenarios. In contrast, our MME-SCI comprehensively covers five languages, enabling systematic assessment of models' understanding and reasoning of scientific concepts across different linguistic contexts.

In terms of modal evaluation dimensions, existing benchmarks have limited coverage. For instance, VisioMath \cite{visiomath} focuses on scenarios with image options, while MATH-Vision \cite{mathvision} centers on image-text hybrid scenarios. However, in practical applications, MLLMs not only process image-text hybrid inputs but also frequently need to handle pure image or pure text inputs. In contrast, our MME-SCI can simultaneously support independent testing for three types of modalities: text-only, image-only, and image-text hybrid. Furthermore, existing benchmarks generally adopt coarse-grained knowledge point classification, whereas MME-SCI annotates each question with fine-grained knowledge points (e.g., ``Trigonometric Functions and Solving Triangle'' in mathematics, ``Regulation of Plant Life Activities'' in biology). This design enables it to more accurately identify models' weaknesses in specific knowledge points compared to existing benchmarks, providing fine-grained diagnostic references for MLLMs.
\section{Experiments}

\subsection{Evaluation details}

We conducted extensive experiments on 16 open-source models and 4 closed-source models. Specifically, the open-source models, which range from 2B to 78B, including Llava-OneVision (7B/72B) \cite{Llava-OneVision}, Qwen2.5VL (3B/7B/72B) \cite{Qwen2.5VL}, Qwen2.5-Omni (7B) \cite{Qwen2.5VLomni}, Ovis2 (8B/16B/34B) \cite{Ovis2}, InternVL3 (2B/8B/78B) \cite{internvl3}, Kimi-VL-A3B-Instruct/Thinking \cite{Kimi-vl}, Phi-4-multimodal-instruct (4.2B) \cite{phi4}, Skywork-R1V-38B \cite{Skyworkr1v}. For the closed-source models, we employed Claude-4-Sonnet \cite{Claude4Sonnet}, Doubao-1.5-thinking-vision-pro \cite{doubao1.5tvp}, Doubao-Seed-1.6 \cite{doubao1.6}, and o4-mini-20250416 \cite{o4mini}

We adopt the `LLM-as-a-Judge' paradigm \cite{llm-as-a-judge,llm-as-a-judge-survey} and introduce different evaluation templates for various languages to assess the models. Unless specified otherwise, we configured the maximum number of new tokens to 8,192, and the temperature was set to 0. 
All experiments are conducted on 8×H20 GPUs. More details can be found in the Appendix.

\subsection{Main Results}

The experimental results on MME-SCI are shown in Table \ref{tab:main_res}, from which the following conclusions can be drawn:

\noindent\textbf{1) A significant gap between open-source and closed-source models.} Compared to advanced closed-source models, the most powerful open-source models (Large group) demonstrates an average accuracy reduction of 13.94\% across six scenarios. Specifically, in the $\boldsymbol{\mathrm{D_{zh}^{img}}}$ scenario, which demands higher visual capabilities of the MLLM, the closed-source models achieved a gain of 125.84\% over the Large group. These results highlight the considerable gap between existing open-source and closed-source models in scientific reasoning tasks, further confirming the challenges posed by MME-SCI to most existing MLLMs.

\noindent\textbf{2) Models with reasoning capabilities have demonstrated significant competitive advantages.} In the context of scientific disciplines, models typically require extensive and complex reasoning, which places higher demands on the model's reasoning capabilities. Compared to models that have not undergone specialized reasoning training, models with reasoning capabilities clearly perform better in such scenarios. For instance, the Thinking version of Kimi-VL-A3B achieves a 48.54\% performance gain over its Instruct version on the $\boldsymbol{\mathrm{D_{zh}}}$, and even achieves an astonishing 72.88\% gain in Image-only scenarios. Additionally, for MLLMs of similar parameter sizes, such as Ovis2-34B and Skywork-R1V-38B, the latter outperforms the former by 4.09\% in average accuracy across six evaluation scenarios.

\noindent\textbf{3) There are significant differences across subjects.} Overall, all models perform noticeably better in the chemistry and biology domains than in mathematics and physics, as the latter require a higher level of reasoning capability. Furthermore, most models perform better in mathematics than in physics, which can be attributed to the fact that physics not only requires complex reasoning but also demands that the model understands the physical laws of the real world \cite{PhyX}. Additionally, there are evident cases of specialization in individual models. For instance, Qwen2.5VL-72B performs only 0.66\% worse than o4-mini in biology, but is 31.34\% worse in mathematics, leading to an overall poorer performance. These findings provide valuable insights for improving future model training, particularly by adjusting the training data composition to enhance performance across multiple aspects.

\noindent\textbf{4) Most models exhibit performance degradation when processing Image-only mode, with open-source models being particularly affected.} Compared to the $\boldsymbol{\mathrm{D_{zh}}}$, the screenshot-based inputs in the $\boldsymbol{\mathrm{D_{zh}^{img}}}$ pose greater challenges to open-source models—specifically, their accuracy in the Small group, Middle group, and Large group drop by 2.79\%, 4.36\%, and 4.29\%, respectively. In contrast, closed-source models, particularly Doubao-1.5-think-vision-pro, achieve a 5.20\% accuracy improvement on the $\boldsymbol{\mathrm{D_{zh}^{img}}}$. This phenomenon indicates that closed-source models, perhaps benefiting from their OCR capabilities, are better able to adapt to image-only tasks, whereas open-source models still face significant challenges under such an evaluation mode. Thus, future work should focus on enhancing the ability of MLLMs to understand screenshot-based inputs, thereby improving their robustness in real-world application scenarios.

\noindent\textbf{5) The visual reasoning ability of the Any2any model degrades.} We specifically evaluated the Any2any model corresponding to Qwen2.5VL-7B, namely Qwen2.5-Omni-7B. The results indicate that, compared to the original vision-language model, its extended Any2any model with additional modalities shows a degradation in visual reasoning ability, with a 4.71\% drop on the $\boldsymbol{\mathrm{D_{zh}^{img}}}$. Therefore, in future model expansions to include more modalities, it is crucial to minimize the degradation of the original capabilities in order to truly achieve an ``omnidirectional'' model.
\section{Further Analysis}

In this section, leveraging the characteristics of our MME-SCI, we have conducted in-depth analytical experiments on contextual information, language consistency, knowledge point differences, and error causes. Additional analytical experiments can be found in the Appendix C.

\subsection{Impacts of in-context learning setting}

We use knowledge point descriptions as context to investigate the impact of prior knowledge on model performance. Specifically, we employ Doubao-Seed-1.6 and Qwen2.5-VL-7/72B as prior knowledge generators to generate both concise and detailed descriptions of knowledge points for each sample in the $\boldsymbol{\mathrm{D_{zh}}}$, with manual verification to ensure that these descriptions do not contain answer information. During the testing phase, the generated prior knowledge points are input into MLLMs as context along with the questions. Results in Tables \ref{tab:simplecot} and \ref{tab:detailedcot} demonstrate that knowledge generated by stronger models can improve the performance of weaker models to a certain extent, while knowledge generated by weaker models exert a negative effect on stronger models, leading to performance degradation.

\begin{table}[!t]
    \centering
    \setlength{\tabcolsep}{4.0pt}
    \begin{tabular}{l|ccc}
        \toprule
        \diagbox{Model}{CoT} & \makecell{Doubao\\-Seed-1.6} & \makecell{Qwen2.5\\VL-7B} & \makecell{Qwen2.5\\VL-72B} \\
        \midrule
        Qwen2.5VL-7B    & 15.51\scriptsize{(+0.30)} & 13.84\scriptsize{(-1.37)} & 14.72\scriptsize{(-0.49)} \\
        Qwen2.5VL-72B   & 22.22\scriptsize{(+0.14)} & 20.12\scriptsize{(-1.96)} & 19.92\scriptsize{(-2.16)} \\
        \bottomrule
    \end{tabular}
    \caption{The performance on $\boldsymbol{\mathrm{D_{zh}}}$ using \textit{concise} pre-knowledge descriptions as context. The baseline of Qwen2.5VL-7B/72B are 15.21\% and 22.08\%, respectively.}
    \label{tab:simplecot}
\end{table}

\begin{table}[!t]
    \centering
    \setlength{\tabcolsep}{4.0pt}
    \begin{tabular}{l|ccc}
        \toprule
        \diagbox{Model}{CoT} & \makecell{Doubao\\-Seed-1.6} & \makecell{Qwen2.5\\VL-7B} & \makecell{Qwen2.5\\VL-72B} \\
        \midrule
        Qwen2.5VL-7B    & 15.90\scriptsize{(+0.69)} & 14.23\scriptsize{(-0.98)} & 14.52\scriptsize{(-0.69)} \\
        Qwen2.5VL-72B   & 22.50\scriptsize{(+0.42)} & 19.53\scriptsize{(-2.55)} & 19.53\scriptsize{(-2.55)} \\
        \bottomrule
    \end{tabular}
    \caption{The performance on $\boldsymbol{\mathrm{D_{zh}}}$ using \textit{detailed} pre-knowledge descriptions as context. The baseline of Qwen2.5VL-7B/72B are 15.21\% and 22.08\%, respectively.}
    \label{tab:detailedcot}
\end{table}

\subsection{Impacts of various languages}

\begin{figure}[!t]
    \centering
    \includegraphics[width=0.99\linewidth]{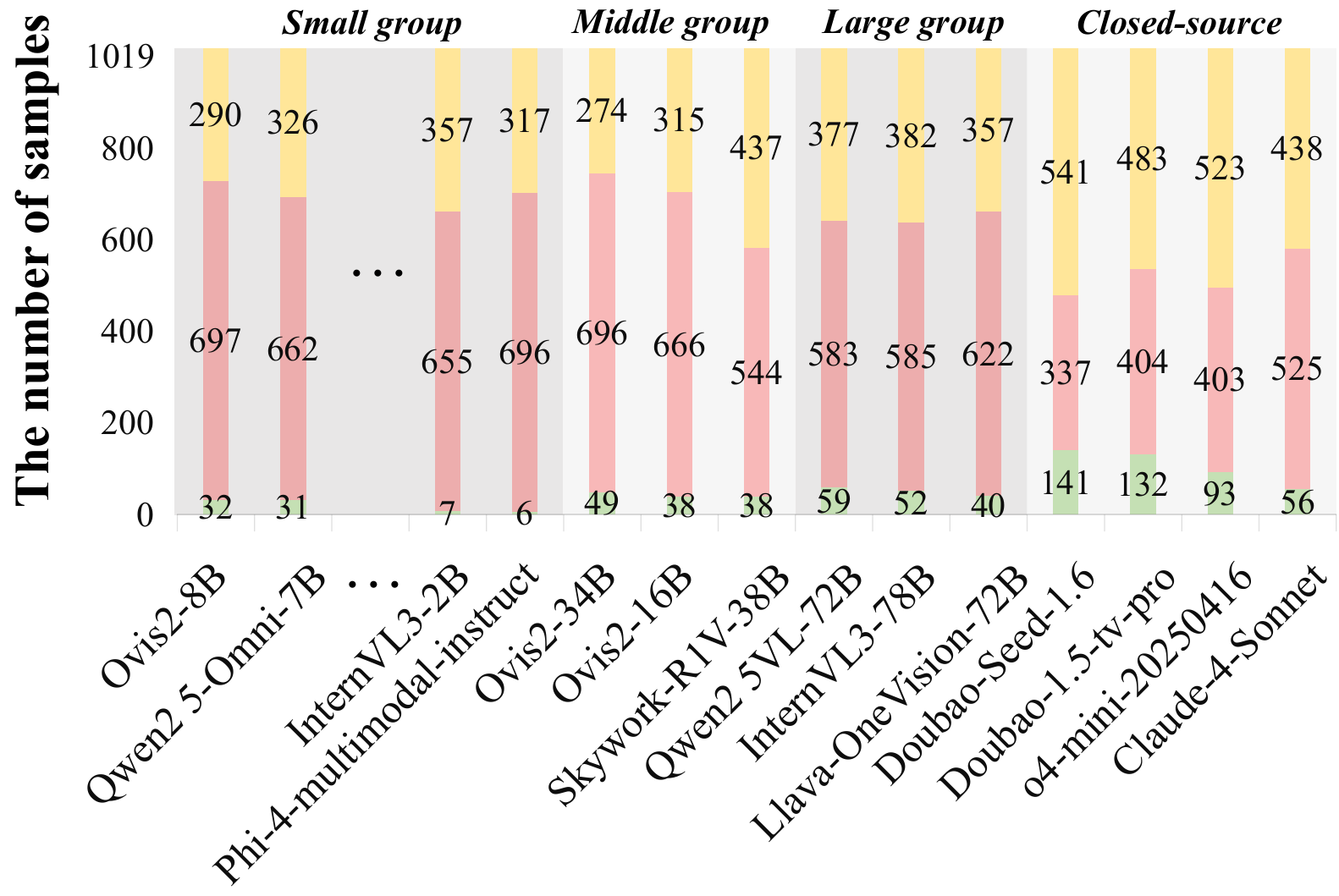}
    \caption{Consistency of responses across five languages. The green section at the bottom shows samples answered correctly consistently across all five languages; the middle red section shows those answered incorrectly consistently across all five. The yellow section at the top is the rest.}
    \label{fig:lan_consis}
\end{figure}

As shown in Figure \ref{fig:lan_consis}, we present the consistency results of responses from different models to the same question across five language scenarios. It can be observed that as model capabilities improve, the number of samples with consistent correct responses shows an increasing trend. This indicates that the enhancement of a model's fundamental capabilities enables it to perform more consistent reasoning in cross-lingual scenarios. However, from an overall data perspective, the MME-SCI dataset contains a total of 1,019 questions. Even the top-performing Doubao-Seed-1.6 only achieves a 13.84\% of linguistically consistent correct responses, while the worst-performing Phi-4 achieves merely a 0.59\% in this regard. This suggests that future MLLMs should focus more on learning truly reasoning abilities rather than merely adapting to specific languages.

\subsection{Impacts of fine-grained knowledge points}

\begin{figure}[!t]
    \centering
    \includegraphics[width=0.99\linewidth]{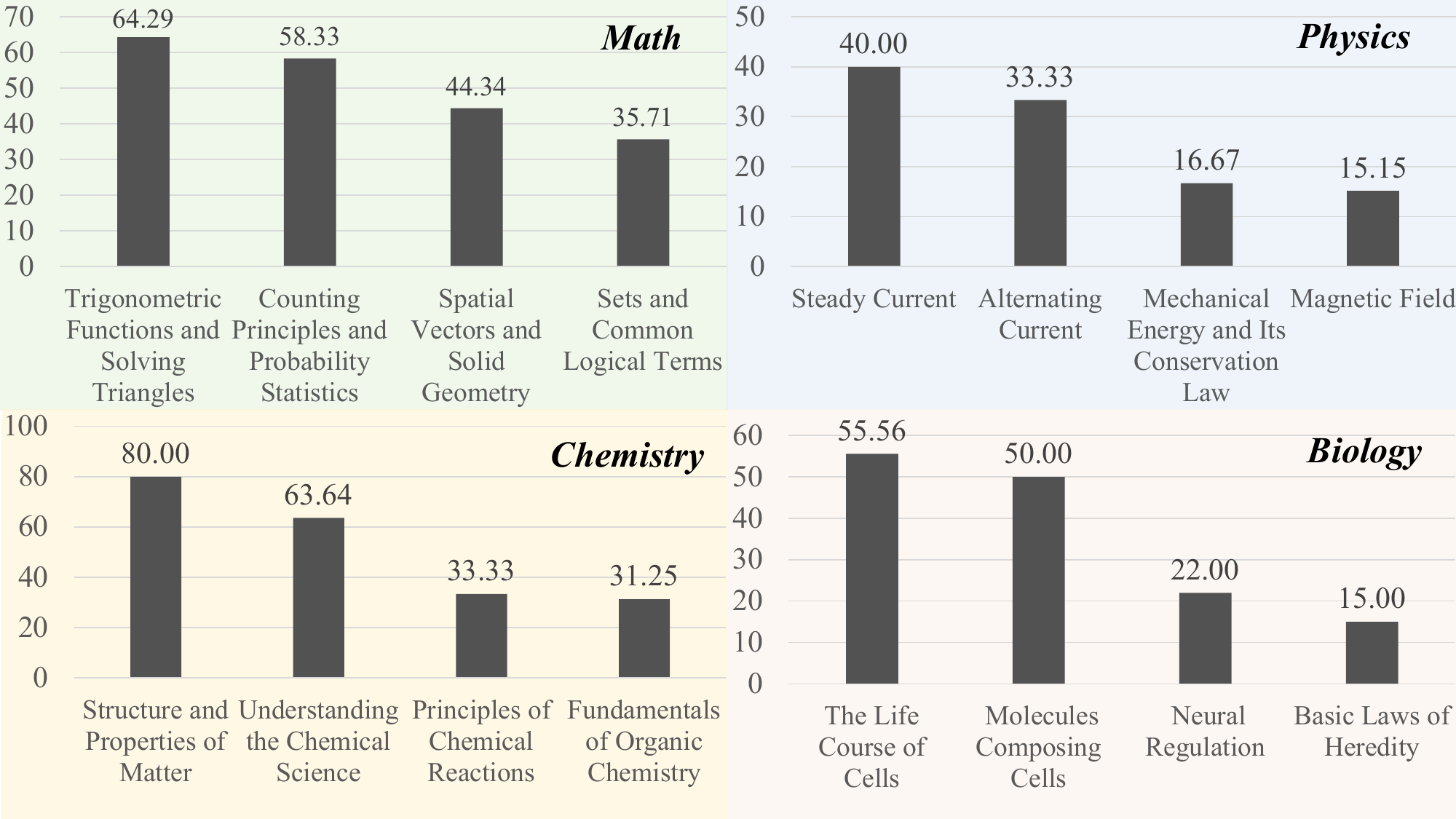}
    \caption{Performance of o4-mini on $\boldsymbol{\mathrm{D_{zh}}}$ with fine-grained knowledge points. We present the top-2 and bottom-2 knowledge points in terms of accuracy across 4 subjects.}
    \label{fig:knowledge_error}
\end{figure}

As illustrated in Figure \ref{fig:knowledge_error}, we report the accuracy of o4-mini across individual Knowledge Points (KPs). The results reveal a significant KP-specific bias of o4-mini in the chemistry subject: while the model attains an accuracy of 80.00\% on its dominant KP, its performance drops to only 31.25\% on the ``Fundamentals of Organic Chemistry''. Such within-subject performance discrepancies directly contribute to the model's underperformance in the overall chemistry subject. Furthermore, the individual KP results in physics indicate that the model performs at a consistently low level in this subject. These observations can provide fine-grained and targeted guidance for subsequent model optimization, such as the introduction of additional physical reasoning training.

\subsection{Error Analysis}

\begin{figure}[!t]
    \centering
    \includegraphics[width=0.99\linewidth]{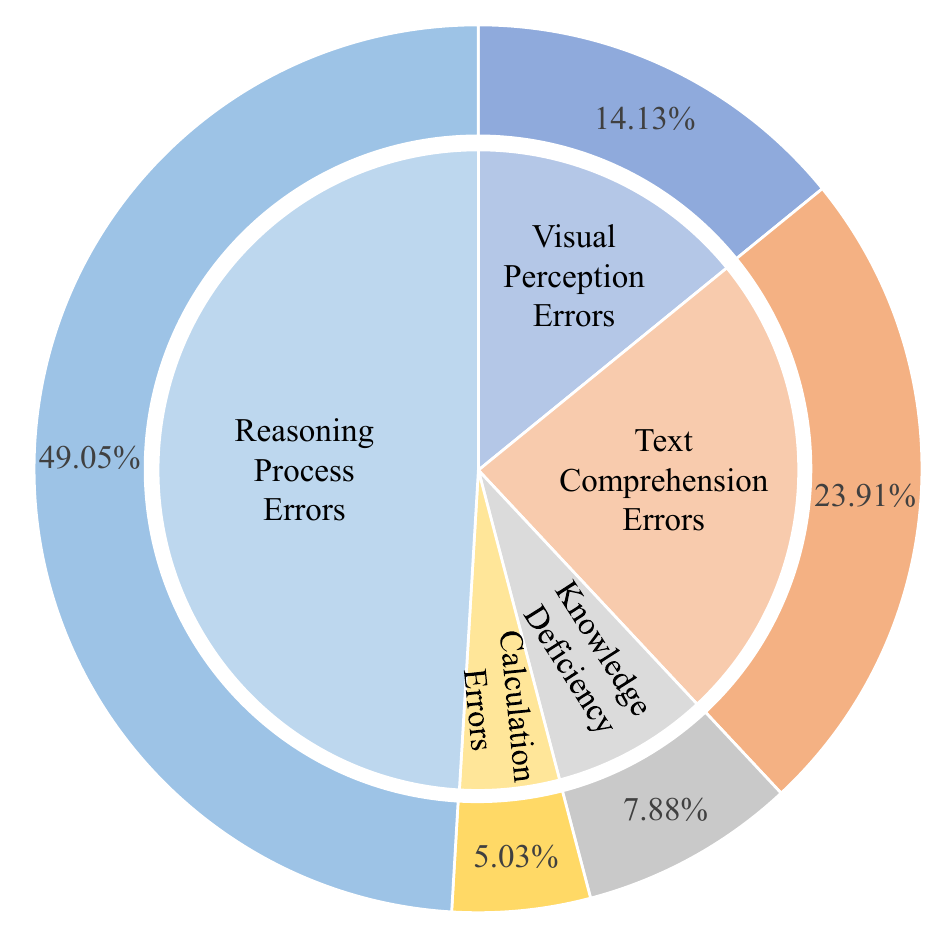}
    \caption{Error distribution on $\boldsymbol{\mathrm{D_{zh}}}$ of Doubao-Seed-1.6. \textbf{Visual perception errors}: response biases caused by incorrect recognition/extraction of image information in questions (e.g., text, symbols, and shapes in the image). \textbf{Text comprehension errors}: errors due to misinterpretation of information, logic, implicit conditions, or explicit instructions in question text. \textbf{Knowledge deficiency}: errors resulting from lack of core domain knowledge (e.g., concepts, theorems) required for the question. \textbf{Calculation errors}: errors in operations, formula substitution, etc., despite correct understanding of question logic and knowledge. \textbf{Reasoning process errors}: errors from flawed logical deduction or causal judgment, despite correct information acquisition and knowledge possession. Furthermore, the erroneous cases can be found in the Appendix D.}
    \label{fig:error_type}
\end{figure}

To thoroughly analyze the defects of MLLMs, we examined the incorrect samples of the Doubao-Seed-1.6 on the $\boldsymbol{\mathrm{D_{zh}}}$ and categorized the error causes, with the results shown in Figure \ref{fig:error_type}. Statistics indicate that reasoning process errors are the most frequent, accounting for 49.05\%, which suggests that the reasoning process is the core link where MLLMs are most prone to mistakes \cite{prm,visualprm,prmbench}. In contrast, calculation errors only account for 5.03\%, indicating that Doubao-Seed-1.6 has already acquired reliable computing capabilities.

\section{Conclusions}

In this paper, we construct MME-SCI, a comprehensive and challenging benchmark characterized by multilingual support, comprehensive modality coverage, multidisciplinary integration, and multi-knowledge point inclusion. We evaluate 20 popular MLLMs on this benchmark, and the results demonstrate that MME-SCI is not only highly challenging but also capable of effectively distinguishing the performance differences among various models. Furthermore, leveraging its inherent characteristics, MME-SCI can help researchers precisely identify the shortcomings of models in language consistency, modal adaptability, reasoning ability, and command of knowledge points. We hope that our MME-SCI will provide directional guidance and research inspiration for the development of MLLMs in the new era.

\bibliography{aaai2026}

\end{document}